\documentclass[11pt]{article}
\usepackage[T1]{fontenc}
\usepackage{textcomp}
\usepackage[utf8]{inputenc}
\usepackage{lmodern,amsmath,amssymb,amsthm,mathtools}
\usepackage[margin=1in]{geometry}
\usepackage{microtype,booktabs,tabularx,enumitem,needspace}
\usepackage[numbers,sort&compress]{natbib}
\usepackage[dvipsnames]{xcolor}
\usepackage{tikz}
\usetikzlibrary{arrows.meta,positioning}
\usepackage[colorlinks=true,linkcolor=blue,citecolor=blue,urlcolor=blue]{hyperref}
\usepackage{aliascnt}
\usepackage{fancyhdr}
\usepackage[skip=\baselineskip]{parskip}
\usepackage{algorithm}
\usepackage{algpseudocode}
\usepackage[capitalise]{cleveref}

\crefname{equation}{}{}
\crefname{figure}{Fig.}{Figs.}
\crefname{section}{Sec.}{Secs.}
\crefname{appendix}{App.}{Apps.}
\crefname{table}{Tab.}{Tabs.}
\newtheorem{theorem}{Theorem}
\crefname{theorem}{Thm.}{Thms.}
\newaliascnt{lemma}{theorem}
\newtheorem{lemma}[lemma]{Lemma}
\aliascntresetthe{lemma}
\crefname{lemma}{Lem.}{Lems.}
\newaliascnt{proposition}{theorem}
\newtheorem{proposition}[proposition]{Proposition}
\aliascntresetthe{proposition}
\crefname{proposition}{Prop.}{Props.}
\newaliascnt{corollary}{theorem}

\aliascntresetthe{corollary}
\crefname{corollary}{Cor.}{Cors.}
\theoremstyle{definition}
\newaliascnt{definition}{theorem}
\newtheorem{definition}[definition]{Definition}
\aliascntresetthe{definition}
\crefname{definition}{Def.}{Defs.}
\theoremstyle{remark}
\newaliascnt{remark}{theorem}

\aliascntresetthe{remark}
\crefname{remark}{Rmk.}{Rmks.}
\crefname{sampler}{Alg.}{Algs.}
\crefname{algorithm}{Alg.}{Algs.}
\newcommand{\R}{\mathbb R}
\newcommand{\E}{\mathbb E}
\newcommand{\N}{\mathcal N}
\newcommand{\KL}{\operatorname{KL}}
\newcommand{\TV}{\operatorname{TV}}
\newcommand{\Var}{\operatorname{Var}}
\newcommand{\Lip}{\operatorname{Lip}}
\newcommand{\osc}{\operatorname{osc}}
\newcommand{\Ent}{\operatorname{Ent}}
\newcommand{\Law}{\operatorname{Law}}
\newcommand{\norm}[1]{\lVert#1\rVert}
\newcommand{\ceil}[1]{\lceil#1\rceil}
\newcommand{\Vpad}{\widetilde V}
\newcommand{\primitive}{\mathsf{NoisySample}}
\newcommand{\decoder}{\mathsf{Decode}}
\renewcommand{\d}{\mathrm{d}}
\newcommand{\e}{\mathrm{e}}
\renewcommand{\top}{\mathrm{T}}

\title{Subpolynomial-Query Proximal Sampling Beyond Log-Concavity}
\author{
Wei Guo\textsuperscript{1}
\quad
Bo Yuan\textsuperscript{1}
\quad
Ye He\textsuperscript{2}
\quad
Jiaming Liang\textsuperscript{3}
\quad
Molei Tao\textsuperscript{1}
\quad
Yongxin Chen\textsuperscript{1}%
\footnote{Correspondence to: \texttt{yongchen@gatech.edu}}
\\[0.5em]
\small
\textsuperscript{1}Georgia Institute of Technology
\quad
\textsuperscript{2}University of Colorado Boulder
\quad
\textsuperscript{3}University of Rochester
}
\date{}
\begin{document}
\maketitle

\begin{abstract}
We study the gradient query complexity of sampling from a density $\pi(x)\propto\e^{-V(x)}$ on $\R^d$, where $V$ has a $\beta$-Lipschitz gradient but need not be convex. Our approach uses the proximal sampler, which alternates Gaussian perturbation with sampling a position conditional on the noisy observation. We implement its observation chain using the high-order noisy conditional sampler of \cite{openai2026}, obtaining improved dimension dependence under functional inequalities. If $\pi$ satisfies a log-Sobolev inequality (LSI) with constant $\alpha$, then for every fixed $\nu>0$, there is a finite constant $C_\nu$ depending only on $\nu$
such that the algorithm returns a sample within total-variation distance $\varepsilon$ using at most
\(C_\nu\kappa(d+\kappa+\varepsilon^{-1}+\|\nabla V(x_0)\|^2/\beta)^\nu\) queries, where $\kappa=\beta/\alpha$ and $x_0$ is an arbitrary reference point for initialization. The analysis controls center-dependent local errors along exact proximal histories and removes the final observation noise by conditional decoding. Under a Poincar\'e inequality (PI), we obtain a corresponding guarantee from a warm start with a known bound on its R\'enyi divergence. An oracle lower bound shows that a stationary point alone, which suffices under LSI, is insufficient under PI.
\end{abstract}

\section{Introduction}\label{sec:introduction}

Sampling from a density $\pi(x)\propto \e^{-V(x)}$ on $\R^d$
is a fundamental task in statistics, scientific computing, and machine
learning. A major line of research studies how the cost of sampling
scales with dimension under strong log-concavity and smoothness.
Underdamped Langevin dynamics and randomized midpoint discretizations
have improved the polynomial dimension dependence of Wasserstein
sampling guarantees~\cite{cheng2018underdamped,shen2019randomized}.
Recently, OpenAI~\cite{openai2026} established a qualitatively different
result: for $C^2$ potentials with a known minimizer and
$I_d\preceq\nabla^2 V\preceq 2I_d$, sampling to total-variation accuracy
$1/10$ requires at most $C_\nu d^\nu$ exact first-order queries for every
fixed $\nu>0$, with unrestricted computation between queries.
Thus, a fixed positive power of the dimension is not intrinsic to
query complexity in this well-conditioned setting.

A parallel line of research replaces strong log-concavity with a
log-Sobolev inequality (LSI) or a Poincar\'e inequality (PI), allowing
the potential to be nonconvex. Analyses of Langevin and proximal
samplers have established polynomial-complexity guarantees under
these weaker assumptions, with appropriate initialization
conditions~\cite{vempala2019rapid,chewi2022analysis,chen2022improved,fan2023improved}; see \cite{chewi2022log} for a comprehensive review.
Under LSI and smoothness assumptions, representative improvements include $d^{1/2}$
dimension dependence for rejection-based proximal
sampling~\cite{fan2023improved} and, more recently, $d^{1/5}$ dependence for the Gaussian cloud
sampler~\cite{chen2026highpicardhmc}.
These guarantees have different accuracy and initialization requirements.
Together with~\cite{openai2026}, they raise a natural question:
\emph{can subpolynomial query complexity in dimension also be achieved
for smooth targets satisfying LSI or PI, with the required initialization
and its cost made explicit?}
We investigate this question through the proximal sampling framework.

Introduced in~\cite{lee2021structured}, the proximal sampler alternates Gaussian perturbation with sampling a position conditional on the noisy observation. This is Gibbs sampling on an augmented space. The conditional draw, called the restricted Gaussian oracle (RGO), adds quadratic regularization to the potential. When the proximal step size is $\tau<1/\beta$, the conditional potential has Hessian bounded below by $(\tau^{-1}-\beta)I_d$ and is therefore strongly convex.
The proximal sampler admits convergence guarantees under LSI and
PI~\cite{chen2022improved}, with efficient RGO implementations based
on approximate rejection sampling~\cite{fan2023improved}. Accelerated variants, including Proximal BPS and reflected proximal dynamics, have also been developed under strong log-concavity~\cite{chen2026accelerated,yuan2026reflected}. Our goal is to improve the dimension dependence of proximal sampling for smooth targets that need not be log-concave.

We build on the local sampling algorithm of~\cite{openai2026} to obtain
query bounds with an arbitrarily small positive dimension exponent.
Its output includes the Gaussian noise needed for the next proximal
step, allowing us to run the chain on noisy positions and recover an
approximate sample from $\pi$ only at the end.
The main difficulty is that the error in each sampling step depends
on the gradient of the potential at the current noisy position.
We use LSI or PI to control these errors along the chain without
assuming global convexity.

\paragraph{Contributions.}
For every fixed $\nu>0$, we obtain the following guarantees with a
finite constant $C_\nu$ depending only on $\nu$.
The algorithm only requires gradient queries to the target potential, and the query complexity holds uniformly over its internal randomness.
\begin{itemize}[leftmargin=*,itemsep=0.2em,topsep=0.3em]
\item
Under LSI and smoothness assumptions, we obtain a sampler with
total-variation error at most $\varepsilon$ using
\[
C_\nu\kappa
\bigl(d+\kappa+\varepsilon^{-1}+G_0\bigr)^\nu
\]
gradient queries for every fixed $\nu>0$ (\cref{thm:lsi}).
Here $\kappa=\beta/\alpha$, where $\beta$ is the smoothness constant
and $\alpha$ is the LSI constant.
The algorithm starts from $\N(x_0,\beta^{-1}I_d)$ at any supplied
point $x_0$, whose quality enters through
$G_0=\norm{\nabla V(x_0)}^2/\beta$.

\item
Under PI and smoothness assumptions, suppose an initial sample
from $\eta_0$ can be generated using at most $Q_{\mathrm{init}}$
gradient queries, with a known bound $D_2(\eta_0\|\pi)\le B$, $B\ge1$.
For every fixed $\nu>0$, we obtain a sampler with total-variation
error at most $\varepsilon$ using at most
\[
Q_{\mathrm{init}}
+C_\nu\kappa\bigl[B+\log(8/\varepsilon)\bigr]
\bigl(d+\kappa+B+\varepsilon^{-1}\bigr)^\nu
\]
gradient queries (\cref{thm:pi}), where $\kappa=\beta/\alpha$
and $\alpha$ is the PI constant.
We also show that a supplied stationary point alone is insufficient
for a uniform query bound under PI, even when it is a strict local
minimum (\cref{thm:obstruction}).

\end{itemize}

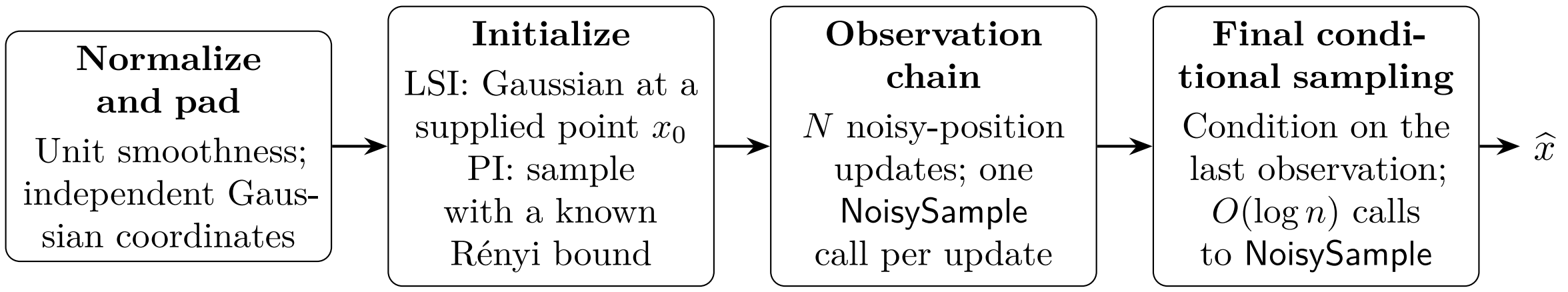
\begin{figure}[t]
\centering
\begin{tikzpicture}[
 stage/.style={
   draw,
   rounded corners,
   align=center,
   text width=2.95cm,
   minimum height=2.2cm,
   font=\small,
   inner sep=4pt
 },
 arr/.style={-{Stealth},thick}
]
\node[stage] (pad) {
  \textbf{Normalize and pad}\\[2pt]
  Unit smoothness; independent Gaussian coordinates
};
\node[stage,right=0.55cm of pad] (init) {
  \textbf{Initialize}\\[2pt]
  LSI: Gaussian at a supplied point $x_0$\\
  PI: sample with a known R\'enyi bound
};
\node[stage,right=0.55cm of init] (chain) {
  \textbf{Observation chain}\\[2pt]
  $N$ noisy-position updates; one $\primitive$ call per update
};
\node[stage,right=0.55cm of chain] (dec) {
  \textbf{Final conditional sampling}\\[2pt]
  Condition on the last observation;
  $O(\log n)$ calls to $\primitive$
};
\node[right=0.4cm of dec] (out) {$\widehat x$};

\draw[arr] (pad) -- (init);
\draw[arr] (init) -- (chain);
\draw[arr] (chain) -- (dec);
\draw[arr] (dec) -- (out);
\end{tikzpicture}
\caption{Structure of \cref{alg:proximal}. The padded target has dimension
$n\ge d$. After the observation updates and final conditional sampling,
we discard the appended coordinates and undo the rescaling to obtain
$\widehat x$.}
\label{fig:pipeline}
\end{figure}

\paragraph{Technical overview.}
The local sampler of~\cite{openai2026} approximates a conditional
position with an independent Gaussian perturbation added. This is
exactly the noise used to form the next proximal observation, so
one call implements an observation update. We therefore run the
chain on noisy positions and, at the end, approximately sample a
position conditioned on the last observation. We call this final
conditional sampling procedure the decoder; \cref{fig:pipeline}
summarizes the algorithm.

The error of each local call depends on the potential gradient at
the current observation. Our comparison argument averages these
errors along the exact proximal chain, where LSI or PI supplies
mixing and gradient-moment bounds. This avoids having to control
gradient moments along the approximate chain. The decoder error is
bounded under the stationary observation law, and
\cref{prop:transfer} combines the mixing, transition, and decoder errors.

\section{Background and main results}\label{sec:main}

\paragraph{Assumption on the target distribution.}
Throughout this paper, we consider the target distribution $\pi(x)\propto\e^{-V(x)}$ on $\R^d$ with $d\ge2$, where $V\in C^2(\R^d)$ is possibly non-convex but has a $\beta$-Lipschitz gradient, so that $-\beta I_d\preceq\nabla^2V\preceq\beta I_d$. The complexity of our algorithms is quantified through the number of gradient queries $x\mapsto\nabla V(x)$, and potential values $V(x)$ are not required for the algorithms.

\paragraph{Notations and definitions.}
For a random variable $X$, $\Law(X)$ denotes its distribution, and for a measurable map $T$, $T_\#\mu:=\Law(T(X))$ for $X\sim\mu$. For probability densities $\mu$ and $\eta$ on $\R^d$, the total-variation distance is $\TV(\mu,\eta)=\sup_A|\mu(A)-\eta(A)|$, the Kullback--Leibler (KL) divergence is $\KL(\mu\Vert\eta)=\int\mu(x)\log\frac{\mu(x)}{\eta(x)}\,\d x$, the $\chi^2$ divergence is $\chi^2(\mu\Vert\eta)=\int\frac{\mu(x)^2}{\eta(x)}\,\d x-1$, and the R\'enyi divergence of order two is $D_2(\mu\Vert\eta)=\log(1+\chi^2(\mu\Vert\eta))$.
Pinsker's inequality gives $2\TV^2\le\KL$, and Jensen's inequality gives $\KL\le\log(1+\chi^2)$.
We write $C$ for a positive constant whose value may change from line to line. It never depends on $d$, $n$, $\kappa$, $\varepsilon$, or the initialization; when it depends on the fixed parameters $(\gamma,K,\omega)$ of $\primitive$ (\cref{prop:primitive}), we mention it explicitly.

\subsection{The proximal sampler}\label{sec:proximal-background}
The proximal sampler \cite{lee2021structured,chen2022improved} augments $x\sim\pi$ with an auxiliary variable $y$ under the joint density $\Gamma_\tau(x,y)\propto\e^{-V(x)-\norm{x-y}^2/(2\tau)}$, where $\tau>0$ is the step size. Its $x$-marginal is $\pi$, and its conditional distributions are
\begin{equation}\label{eq:original-proximal}
 y\mid x\sim\N(x,\tau I_d),
 \qquad
 x\mid y\sim q_y^\tau(x)\propto\e^{-V(x)-\norm{x-y}^2/(2\tau)}.
\end{equation}
An exact draw from $q_y^\tau$ is called a restricted Gaussian oracle (RGO) draw. The proximal sampler is the Gibbs chain that draws $y_k\sim\N(x_k,\tau I_d)$ and then $x_{k+1}\sim q_{y_k}^\tau$. It preserves $\Gamma_\tau$ for every $\tau$, so the \emph{position chain} $(x_k)$ has stationary law $\pi$, and the \emph{observation chain} $(y_k)$ has stationary law $\pi_\tau:=\pi*\N(0,\tau I_d)$. Since $\nabla^2V\succeq-\beta I_d$, every $q_y^\tau$ is strongly log-concave when $\tau<1/\beta$, even if $\pi$ is not. Smaller step sizes make the RGO easier but slow the mixing of the chain, as we will see in \cref{prop:ideal-convergence}.

\subsection{Functional inequalities}
\begin{definition}[Functional inequalities]
Let $\Ent_\pi(g):=\E_\pi[g\log\frac{g}{\E_\pi g}]$ for $g\ge0$.
We say that $\pi$ satisfies an $\alpha$-logarithmic Sobolev inequality (LSI) or an $\alpha$-Poincar\'e inequality (PI) if, respectively, the following inequalities hold for all smooth $f$:
\begin{align}
 \Ent_\pi(f^2) &\le\frac2\alpha\E_\pi\norm{\nabla f}^2, \tag{LSI}\label{eq:lsi}\\
 \Var_\pi f&\le\frac1\alpha\E_\pi\norm{\nabla f}^2. \tag{PI}\label{eq:pi}
\end{align}
Note that $\alpha$-strongly log-concave targets satisfy $\alpha$-LSI, and $\alpha$-LSI implies $\alpha$-PI \cite{bakry2014analysis}.
\end{definition}

Unlike strong log-concavity, both inequalities are stable under bounded perturbations of $V$ \cite{bakry2014analysis}, so they allow saddle points and multiple local minima (see the example at the end of \cref{sec:lsi-proof}). A larger $\alpha$ means the distribution is better concentrated and typically implies faster mixing, and $\kappa:=\beta/\alpha$ plays the role of a condition number as in the strongly log-concave case.
Valid constants always satisfy $\alpha\le\beta$ (\cref{lem:alpha-beta}), so $\kappa\ge1$.

\begin{proposition}[Convergence of the ideal proximal sampler]\label{prop:ideal-convergence}
Fix $\tau>0$ and run \cref{eq:original-proximal} with exact conditional draws. Let $\rho_k=\Law(x_k)$ and assume the relevant initial divergence below is finite. For every integer $k\ge0$,
\begin{align}
 \KL(\rho_k\Vert\pi)&\le(1+\alpha\tau)^{-2k}\KL(\rho_0\Vert\pi),
 &&\text{if $\pi$ satisfies $\alpha$-LSI},\\
 \chi^2(\rho_k\Vert\pi)&\le(1+\alpha\tau)^{-2k}\chi^2(\rho_0\Vert\pi),
 &&\text{if $\pi$ satisfies $\alpha$-PI}.
\end{align}
The same respective upper bounds hold for the divergences of the observation law $\sigma_k=\rho_k*\N(0,\tau I_d)=\Law(y_k)$ from $\pi_\tau$.
\end{proposition}
\begin{proof}
The position bounds follow from \cite[Thms.~3 and~4]{chen2022improved}. Applying the same Gaussian convolution kernel to $\rho_k$ and $\pi$ contracts both divergences by data-processing inequalities, giving the observation bounds.
\end{proof}
These bounds require neither global convexity nor $\tau<1/\beta$ for the ideal chain, and they furthermore imply geometric convergence in TV with factor $(1+\alpha\tau)^{-k}$. With a step size $\tau$ of order $1/\beta$, about $\widetilde{O}(\kappa)$ iterations (up to a logarithmic factor in the initial divergence and the accuracy threshold) therefore suffice for the desired accuracy.

\subsection{Sampling guarantees}
We now state the main theorems of the complexity of the proposed algorithm, \cref{alg:proximal}.
\begin{theorem}[Gaussian-start sampling under LSI]\label{thm:lsi}
Suppose $\pi$ satisfies $\alpha$-LSI.
For every fixed $\nu>0$, there is a finite $C_\nu$ such that the following holds. Let $x_0\in\R^d$ be any point provided as an input to the algorithm, and set $G_0=\norm{\nabla V(x_0)}^2/\beta$. For every $0<\varepsilon<1/4$, \cref{alg:proximal} returns a sample whose law $\widehat\pi$ satisfies $\TV(\widehat\pi,\pi)\le\varepsilon$ using at most
\begin{equation}\label{eq:lsi-main}
 C_\nu\kappa(d+\kappa+\varepsilon^{-1}+G_0)^\nu
\end{equation}
gradient queries.
\end{theorem}

\begin{theorem}[Finite-R\'enyi-start sampling under PI]\label{thm:pi}
Suppose $\pi$ satisfies $\alpha$-PI.
For every fixed $\nu>0$, there is a finite $C_\nu$ such that the following holds. Assume an initial sample $x_{\rm init}\sim\eta_0$ can be generated using only gradient access to $V$, with deterministic query cost at most $Q_{\rm init}$, and a known $B\ge1$ satisfies $D_2(\eta_0\Vert\pi)\le B$. For every $0<\varepsilon<1/4$, \cref{alg:proximal} returns a sample whose law $\widehat\pi$ satisfies $\TV(\widehat\pi,\pi)\le\varepsilon$ using at most
\begin{equation}\label{eq:pi-main}
 Q_{\rm init}+C_\nu\kappa[B+\log(8/\varepsilon)]
 (d+\kappa+B+\varepsilon^{-1})^\nu
\end{equation}
gradient queries.
\end{theorem}

\paragraph{Interpretation.}
Both theorems run \cref{alg:proximal} and differ only in its initialization. Under LSI, $x_0$ is an arbitrary input point; the cost of finding it is not counted, and its quality enters only through $G_0$, which vanishes at any stationary point, including a saddle point. Under PI, the initial sampler and the bound $B$ are inputs, and the cost $Q_{\rm init}$ of the initial sample is charged to the total.
The prefactor $\kappa$ reflects the roughly $\kappa$ iterations needed by the exact proximal sampler (\cref{prop:ideal-convergence}), and all other parameters enter through a power $\nu>0$ that can be taken arbitrarily small, with the algorithm and $C_\nu$ depending on $\nu$. Thus, for fixed $\kappa$, $\varepsilon$, and $G_0$ or $B$, the number of gradient queries grows more slowly than any positive power of the dimension.

\section{Implementation of the proximal sampler}\label{sec:construction}\label{sec:localization}\label{sec:inputs}
The proximal sampler of \cref{sec:proximal-background} needs an exact draw from the RGO $q_y^\tau$ at every step. Rather than implementing these draws directly, our implementation rests on two observations.

First, the Gaussian noise already present in the proximal sampler
allows us to work with a smoothed conditional distribution.
Given an observation $y$, the next observation is
$y'=x'+\sqrt{\tau}\,G$, where $x'\sim q_y^\tau$ and
$G\sim\N(0,I_d)$ are independent.
Writing $x'=y+\sqrt{\tau}\,Z$, we have $y'=y+\sqrt{\tau}(Z+G)$, where $Z$ follows
\[
\operatorname{Law}(Z)(v)
\propto
\e^{-\frac{\norm{v}^2}{2}-V(y+\sqrt{\tau}\,v)}.
\]
Thus, updating the observation requires a sample of $Z+G$.
The additional Gaussian noise smooths the distribution of $Z$.
The construction of~\cite{openai2026} combines this smoothing with
high-order approximation to implement the noisy conditional draw
using few gradient queries.
This is central to obtaining an arbitrarily small dimension exponent.

Second, we observe that the same sampling construction can extend
 beyond strong log-concavity to targets satisfying LSI or PI.
These weaker assumptions still provide enough control over the chain
to establish convergence and bound the accumulated sampling errors.
The two settings require different initialization conditions and analysis, which we make explicit below.

In the following, we first state the guarantee of $\primitive$
from~\citep{openai2026} in \cref{sec:primitive}.
We then describe how to rescale the target and append independent
Gaussian coordinates (\cref{sec:padding}), and how to sample a
position conditioned on the final noisy observation
(\cref{sec:decoder}).
The complete procedure is given in \cref{alg:proximal},
and \cref{prop:transfer} bounds its total approximation error.
Throughout, hats denote approximate kernels and probability laws.

\subsection{The noisy local sampling algorithm}\label{sec:primitive}

For a perturbation $F\in C^2(\R^n)$, center $c\in\R^n$, and scale $0<r\le1$, define
\begin{equation}\label{eq:conditional}
 \rho^F_{c,r}(v)\propto\e^{-\norm v^2/2-F(c+rv)}.
\end{equation}
With $n=d$, $F=F_y$, $c=0$, and $r=1$, this is the law of $Z$ above after the transformation of the RGO target $q_y^\tau$. Given $c$, $r$, and access to $\nabla F$, the algorithm $\primitive$ approximates $Z+G$ for independent $Z\sim\rho^F_{c,r}$ and $G\sim\N(0,I_n)$ without exact sample of $Z$.
Its construction is explained in \cref{app:local-mechanism}.
The next result establishes the convergence guarantee of $\primitive$ and the required gradient query complexity.

\begin{proposition}[Noisy local sampling guarantee]\label{prop:primitive}
Fix a localization exponent $\gamma\in(0,1)$, an approximation order $K\ge2$, and a query-overhead exponent $\omega>0$ such that
\begin{equation}\label{eq:source-gate}
 40(K+2)\omega<1/4.
\end{equation}
There are a constant $C$ and a dimension threshold, both depending only on $(\gamma,K,\omega)$, and a bound $A_n\le C(1+\log n)^C$, with the following property. For every $n$ above the threshold, $\delta=n^{-\gamma}$, $\Lip(\nabla F)\le\delta$, $c\in\R^n$, and $0<r\le1$, the algorithm $\primitive(F,c,r)$ uses at most $A_n\delta^{-\omega}$ gradient queries to $F$ and satisfies
\begin{equation}\label{eq:source-error}
 \TV(\Law(\primitive(F,c,r)),\rho^F_{c,r}*\N(0,I_n))
 \le A_n\delta^K(\sqrt n+\norm{\nabla F(c)}).
\end{equation}
The constants are uniform in $F,c,r$.
\end{proposition}
This guarantee follows from \cite[Thm.~2.1]{openai2026}. We suppress the fixed parameters $(\gamma,K,\omega)$ in the notation $\primitive(F,c,r)$.

\subsection{Normalization and Gaussian padding}\label{sec:padding}
We now construct the padded target described at the beginning of this section.
Let $a=x_0$ under LSI and $a=0$ under PI. For $w=(z,\xi)\in\R^d\times\R^{n-d}$, define the forward and output maps
\begin{equation}\label{eq:coordinate-maps}
 T_a(x)=\sqrt\beta(x-a),\qquad H_a(w)=a+z/\sqrt\beta,
\end{equation}
and the padded potential and target
\begin{equation}\label{eq:padded}
 \Vpad(w)=V(a+z/\sqrt\beta)+\norm\xi^2/2,\qquad
 \Pi=(T_a)_\#\pi\otimes\N(0,I_{n-d})\propto\e^{-\Vpad}.
\end{equation}
Then $H_{a\#}\Pi=\pi$, the potential $\Vpad$ is 1-smooth with LSI or PI constant $1/\kappa$, and each gradient of $\Vpad$ costs one gradient of $V$ (\cref{lem:normalization}). In these coordinates the step size becomes $h=\beta\tau$.

\paragraph{Proximal chain on the padded target.}
The two conditionals of \cref{eq:original-proximal}, for $\Pi$ and step size $h$, are the kernels
\[
 A_h(y\mid w)=(2\pi h)^{-n/2}\e^{-\norm{y-w}^2/(2h)},\qquad
 R_h(w\mid y)\propto\e^{-\Vpad(w)-\norm{w-y}^2/(2h)}.
\]
For a law $\mu$, write $\mu A_h(y)=\int\mu(w)A_h(y\mid w)\,\d w$, similarly for $R_h$, and compose kernels from left to right by convention. The observation chain $y\to w'\to y'$ then has kernel $P_h=R_hA_h$. With $\Pi_h:=\Pi A_h=\Pi*\N(0,hI_n)$, we have
\[
 \Pi_hP_h=\Pi_h,\qquad \Pi_hR_h=\Pi,
\]
i.e., the observation chain is stationary at $\Pi_h$, and a position drawn given a stationary observation is an exact sample of $\Pi$, the step the decoder in \cref{sec:decoder} approximates. Let $X_j\sim\lambda_j$ and $Y_j\sim\mu_j$ be the positions and observations of the exact chain, so that $\mu_j=\lambda_jA_h=\mu_0P_h^j$.

\subsection{Implementing one observation step}\label{sec:observation-implementation}
Let $F_y(v)=\Vpad(y+\sqrt h v)$. By the argument at the beginning of this section, $P_h(\,\cdot\mid y)$ is the law of $y+\sqrt h(Z+G)$ for independent $Z\sim\rho^{F_y}_{0,1}$ and $G\sim\N(0,I_n)$. The implemented kernel $\widehat P_h(\,\cdot\mid y)$ replaces $Z+G$ by the output of $\primitive$:
\[
 \widehat Y'=y+\sqrt h\,\primitive(F_y,0,1).
\]
The gradient $g_y(v)=\nabla F_y(v)=\sqrt h\nabla\Vpad(y+\sqrt h v)$ is $h$-Lipschitz, and in \cref{alg:proximal} we have $h=\delta/2\le\delta$ and $n$ above the threshold of \cref{prop:primitive}. Since TV is unchanged by the common affine map $u\mapsto y+\sqrt h\,u$, \cref{eq:source-error} with $c=0$, $r=1$, and $\nabla F_y(0)=\sqrt h\nabla\Vpad(y)$ gives
\begin{equation}\label{eq:row-error}
 \TV(\widehat P_h(\,\cdot\mid y),P_h(\,\cdot\mid y))
 \le A_n\delta^K(\sqrt n+\sqrt h\norm{\nabla \Vpad(y)}).
\end{equation}

\subsection{The final decoder}\label{sec:decoder}
After $N$ steps the observation chain is close to $\Pi_h$, not $\Pi$. A position sample requires one draw from $R_h(\,\cdot\mid y)$ at $y=Y_N$, that is, $y+\sqrt hZ$ with $Z\sim\rho^{F_y}_{0,1}$, but $\primitive$ only returns the noisy variable $Z+G$. The decoder closes this gap with the following identity.

\begin{lemma}[Variance halving \cite{openai2026}]\label{lem:halving}
Let $Z\sim\rho^F_{c,r}$ and $G\sim\N(0,I_n)$ be independent, and set $U=Z+G$. Then
\begin{equation}\label{eq:halving}
 (Z \mid U=u)\stackrel{\rm law}=u/2+Z'/\sqrt2,
 \qquad\text{where}\quad Z'\sim\rho^F_{c+ru/2,\,r/\sqrt2}.
\end{equation}
\end{lemma}

Equivalently, given one noisy observation $u$, the variable $c+rZ$ has the law of $c'+r'Z'$ for the same family with center $c'=c+ru/2$ and scale $r'=r/\sqrt2$. Repeating this with fresh calls to $\primitive$ shrinks the scale geometrically, until the remaining conditional is nearly Gaussian and can be sampled directly. For fixed $y$, the decoder runs as follows.
\begin{enumerate}[leftmargin=*,itemsep=2pt,topsep=4pt]
\item Set $c_0=0$, $r_0=1$, and $J=\ceil{4\log_2n}$. For $j=0,\ldots,J-1$, draw $U_j=\primitive(F_y,c_j,r_j)$ and set $c_{j+1}=c_j+r_jU_j/2$ and $r_{j+1}=r_j/\sqrt2$.
\item Now $r_J\le n^{-2}$, so $\rho^{F_y}_{c_J,r_J}$ is close to a Gaussian with identity covariance centered at its mode, which is the fixed point of $v\mapsto-r_Jg_y(c_J+r_Jv)$. Approximate the mode by $m=\ceil{\log n}$ fixed-point steps,
\begin{equation}\label{eq:fixed-point}
 v^{(0)}=0,\qquad v^{(k+1)}=-r_Jg_y(c_J+r_Jv^{(k)}).
\end{equation}
\item Draw a fresh $G\sim\N(0,I_n)$ and return
\begin{equation}\label{eq:decoded-output}
 \decoder(y)=y+\sqrt h\{c_J+r_J(v^{(m)}+G)\}.
\end{equation}
\end{enumerate}
Write $\widehat R_h$ for the kernel $y\mapsto\Law(\decoder(y))$. With exact draws in place of $\primitive$ and of the final Gaussian approximation, \cref{lem:halving} shows that $\decoder(y)\sim R_h(\,\cdot\mid y)$ exactly.
For an approximate implementation, we quantify the error and query complexity in the following proposition.

\begin{proposition}[Decoder accuracy]\label{prop:decoder}
For every smooth normalizable $\Vpad$ as in \cref{lem:score}, the decoder satisfies
\begin{equation}\label{eq:decoder-error}
 \TV(\Pi_h\widehat R_h,\Pi)
 \le CJA_n\delta^K\sqrt n+C\delta n^{-7/2}.
\end{equation}
Its deterministic query cost is at most $JA_n\delta^{-\omega}+m$.
\end{proposition}

The proof is in \cref{app:decoder}. Its global input is the stationary score identity $\E_\Pi\norm{\nabla \Vpad}^2\le n$. It does not require global convexity or proximity to a minimizer.

\paragraph{The outer algorithm}
\cref{alg:proximal} assembles the pieces of this section: padding, $N$ implemented observation steps, and the decoder. The dimension $n$ and the number of steps $N$ are explicit functions of the inputs; they involve a fixed threshold $n_*(\nu)$ depending only on $\nu$ (\cref{app:parameters}).

\begin{algorithm}[t]
\caption{Subpolynomial-Query  Proximal sampling}
\label{alg:proximal}
\small
\begin{algorithmic}[1]

\Require Normalized and padded potential $\Vpad$ on $\R^n$;
initial position $X_0$; step size $h$; number of iterations $N$;
original dimension $d$; shift $a\in\R^d$ and scale $\beta>0$.

\Statex \textbf{Observation updates}
\State Draw $G\sim\N(0,I_n)$
and set $Y_0\gets X_0+\sqrt{h}\,G$.
\For{$j=0,\ldots,N-1$}
    \State Define $F_{Y_j}(v)=\Vpad(Y_j+\sqrt{h}\,v)$.
    \State $U_j\gets\primitive(F_{Y_j},0,1)$.
    \State $Y_{j+1}\gets Y_j+\sqrt{h}\,U_j$.
\EndFor

\Statex
\Statex \textbf{Sample a position conditioned on $Y_N$}
\State Define $F(v)=\Vpad(Y_N+\sqrt{h}\,v)$.
\State Set $c\gets 0$, $r\gets 1$,
and $J\gets\lceil 4\log_2 n\rceil$.
\For{$\ell=1,\ldots,J$}
    \State $U\gets\primitive(F,c,r)$.
    \State $c\gets c+\frac{r}{2}U$.
    \State $r\gets r/\sqrt{2}$.
\EndFor

\Statex
\Statex \textbf{Approximate the remaining conditional by a Gaussian}
\State Set $v\gets 0$ and $m\gets\lceil\log n\rceil$.
\For{$k=1,\ldots,m$}
    \State
    $v\gets-r\sqrt{h}\,
    \nabla\Vpad\!\left(Y_N+\sqrt{h}(c+rv)\right)$.
    \Comment{Approximate the conditional mode}
\EndFor
\State Draw a fresh $G\sim\N(0,I_n)$.
\State $\widehat W\gets
Y_N+\sqrt{h}\bigl[c+r(v+G)\bigr]$.

\Statex
\Statex \textbf{Return to the original coordinates}
\State $\widehat x\gets
a+\beta^{-1/2}(\widehat W_1,\ldots,\widehat W_d)^\top$.
\State \Return $\widehat x$.

\end{algorithmic}
\end{algorithm}

\paragraph{Query accounting.}
All random draws in \cref{alg:proximal} are independent, and $n$, $N$, and all loop lengths are determined by the inputs, so the number of gradient queries has a deterministic upper bound. The algorithm uses only gradients of $V$. Taking $n\ge n_*(\nu)$ also avoids a low-dimensional fallback in \cite{openai2026} that would require function values. Under PI, the initial sampler is likewise assumed to use only gradients.

\subsection{Accumulation of approximation errors}
\cref{alg:proximal} has three sources of error: the exact observation chain is run for only $N$ steps, each step of the observation chain uses $\primitive$ with error \cref{eq:row-error}, and the decoder is approximate. The following proposition combines them. We remark that this result does not require functional inequalities; those will enter only through the two inputs $\varepsilon_{\rm mix}$ and $M_N$, which concern the exact chain.

\begin{proposition}[From exact histories to implemented samples]\label{prop:transfer}
Let $\mu_j=\mu_0P_h^j$ be the exact observation laws and $\widehat\mu_N=\mu_0\widehat P_h^N$ the implemented one. Suppose \cref{eq:row-error} holds, $\TV(\mu_N,\Pi_h)\le\varepsilon_{\rm mix}$, and $\TV(\Pi_h\widehat R_h,\Pi)\le\varepsilon_{\rm dec}$. Define
\begin{equation}
 M_N=\sum_{j=0}^{N-1}\E_{\mu_j}[\sqrt n+\sqrt h\norm{\nabla \Vpad(Y)}].
\end{equation}
Then
\begin{equation}\label{eq:four-errors}
 \TV(\widehat\mu_N\widehat R_h,\Pi)
 \le\varepsilon_{\rm mix}+A_n\delta^K M_N+\varepsilon_{\rm dec}.
\end{equation}
Moreover, the number $Q_{\rm post}$ of gradient queries made \underline{after} initialization satisfies the deterministic bound
\begin{equation}\label{eq:query-general}
 Q_{\rm post}\le(N+J)A_n\delta^{-\omega}+m.
\end{equation}
\end{proposition}
\begin{proof}
For $0\le k\le N$, let $\theta_k=\mu_0P_h^k\widehat P_h^{N-k}$ be the law after $k$ exact steps followed by $N-k$ implemented steps, so that $\theta_0=\widehat\mu_N$ and $\theta_N=\mu_N$. The laws $\theta_k$ and $\theta_{k+1}$ differ only in step $k+1$, which starts from $\mu_k$, and TV does not increase under the common remaining kernel $\widehat P_h^{N-k-1}$ by the data-processing inequality. By the triangle inequality, summing over $k$ and using \cref{eq:row-error}, we have
\begin{equation}
 \TV(\widehat\mu_N,\mu_N)
 \le\sum_{k=0}^{N-1}\E_{\mu_k}
       \TV(\widehat P_h(\,\cdot\mid Y),P_h(\,\cdot\mid Y))
 \le A_n\delta^K M_N.
\end{equation}
Applying the same decoder kernel $\widehat R_h$ to $\widehat\mu_N$, $\mu_N$, and $\Pi_h$ and using the triangle inequality,
\begin{equation}
 \TV(\widehat\mu_N\widehat R_h,\Pi)
 \le\TV(\widehat\mu_N,\mu_N)+\TV(\mu_N,\Pi_h)+\TV(\Pi_h\widehat R_h,\Pi),
\end{equation}
which gives \cref{eq:four-errors}. There are $N+J$ calls to $\primitive$ and $m$ fixed-point steps, which proves \cref{eq:query-general}.
\end{proof}

In summary, \cref{prop:transfer} reduces the analysis of \cref{alg:proximal} to two properties of the exact chain: its mixing error $\varepsilon_{\rm mix}$ and the score budget $M_N$, which averages the gradient under the exact laws $\mu_j$ rather than the implemented ones. In the next two sections, we bound both quantities under LSI (\cref{sec:lsi-proof}) and under PI (\cref{sec:pi-proof}).

\section{Sampling under LSI}\label{sec:lsi-proof}
Under LSI, both $\varepsilon_{\rm mix}$ and $M_N$ follow from a bound on the KL divergence of the Gaussian initialization (\cref{lem:gaussian-kl}). This divergence decays geometrically along the exact chain (\cref{prop:ideal-convergence}), which controls $\varepsilon_{\rm mix}$; since it never increases, Talagrand's transport inequality also bounds the gradient moments in $M_N$ (\cref{lem:lsi-moments}).

\begin{proposition}[Exact-chain inputs under LSI]\label{prop:lsi-inputs}
Let $n\ge\max\{d,\kappa,G_0,\varepsilon^{-1}\}$, $\delta\le1/2$, and $h=\delta/2$, and let $N$ be as in \cref{eq:N-lsi}. For the exact chain started from $\lambda_0=\N(0,I_n)$,
\[
 \varepsilon_{\rm mix}\le\varepsilon/4,\qquad
 M_N\le CN\kappa\sqrt n,\qquad
 N\le C\kappa\delta^{-1}\log n.
\]
\end{proposition}

Substituting these bounds into \cref{prop:transfer}, the implementation error is at most $C\kappa^2A_n\delta^{K-1}\sqrt n\log n$ and the algorithm makes at most $C\kappa A_n(\log n)\,n^{\gamma(1+\omega)}$ queries. Choosing $\gamma$ small makes the query exponent less than $\nu$, and choosing $K$ large relative to $1/\gamma$ then makes the implementation and decoder errors at most $\varepsilon/4$ each. We detail the explicit choices in \cref{app:parameters}. Mapping the output back with $H_a$ gives \cref{thm:lsi}. Full proof can be found in \cref{app:lsi-proofs}. Note that the initial divergence can be of order $\kappa n$, but it enters $N$ only logarithmically.

\paragraph{A non-log-concave example.}
The potential $V_d(x)=\norm x^2/2+2\cos x_1$ is $3$-smooth, has a negative Hessian eigenvalue at zero, and has LSI constant at least $\e^{-4}$ by bounded perturbation (\cref{app:example}). Since $\nabla V_d(0)=0$, \cref{thm:lsi} applies with $x_0=0$ and $G_0=0$. As $\kappa\le3\e^4$ is a constant, it gives $O_\nu((d+\varepsilon^{-1})^\nu)$ gradient queries for every fixed $\nu>0$. For fixed accuracy, this improves the dimension dependence of the $\widetilde O(\kappa\sqrt d)=\widetilde O(\sqrt d)$ bound in \cite{fan2023improved} and the $\widetilde O(\kappa d^{1/5})=\widetilde O(d^{1/5})$ bound in \cite{chen2026highpicardhmc}, whose condition $G_0\le d/\kappa$ holds here.

\section{Sampling under PI and initialization}\label{sec:pi-proof}
\subsection{Proof outline}
We outline the proof of \cref{thm:pi}. As under LSI, it supplies the two exact-chain inputs $\varepsilon_{\rm mix}$ and $M_N$ of \cref{prop:transfer}, but two ingredients change. First, PI does not bound the divergence of a simple Gaussian start, so the initial law $\eta_0$ and its bound $D_2(\eta_0\Vert\pi)\le B$ are inputs. Under PI the exact chain contracts $\chi^2$ (\cref{prop:ideal-convergence}), and $D_2\le B$ allows $\chi^2$ as large as $\e^B-1$, so the number of steps grows linearly in $B$ rather than logarithmically. Second, PI alone does not imply the quadratic transport inequality
used in the LSI argument. It instead gives exponential concentration of Lipschitz functions (\cref{lem:pi-mgf}), which, by a change of measure, bounds the first gradient moment under any law within KL divergence $B$ of $\Pi$ (\cref{lem:pi-score}). A first moment is all that $M_N$ requires. The result is as follows:

\begin{proposition}[Exact-chain inputs under PI]\label{prop:pi-inputs}
Let $n\ge\max\{d,\kappa(B+1)^2\}$, $\delta\le1/2$, and $h=\delta/2$, let $N$ be as in \cref{eq:N-pi}, and set $H=B+\log(8/\varepsilon)$. For the exact chain started from $\lambda_0=(T_0)_\#\eta_0\otimes\N(0,I_{n-d})$,
\[
 \varepsilon_{\rm mix}\le\varepsilon/4,\qquad
 M_N\le CN\sqrt n,\qquad
 N\le C\kappa H\delta^{-1}.
\]
\end{proposition}

Substituting these bounds into \cref{prop:transfer}, the implementation error is at most $C\kappa HA_n\delta^{K-1}\sqrt n$, and the algorithm makes at most $Q_{\rm init}+C\kappa HA_n\,n^{\gamma(1+\omega)}$ queries. The parameters are then chosen as under LSI (\cref{app:parameters}), and mapping the output back with $H_0$ gives \cref{thm:pi}; \cref{app:pi-proofs} contains the full proof.

\subsection{Initialization under PI}\label{sec:obstruction}
\cref{thm:pi} takes the initial law $\eta_0$ and its bound $D_2(\eta_0\Vert\pi)\le B$ as inputs. What information about $\pi$ suffices to construct them? We now demonstrate that a global minimizer does, but an arbitrary stationary point does not.

\paragraph{A global minimizer suffices.}
If $x_*$ is a global minimizer of $V$, the Gaussian $q_0=\N(x_*,(2\beta)^{-1}I_d)$ satisfies $D_2(q_0\Vert\pi)\le\frac d2\log(2\e\kappa)$ (see \cref{prop:gaussian-pi}, which also covers approximate minimizers). \cref{thm:pi} thus applies with $Q_{\rm init}=0$ and $B=\max\{1,\tfrac d2\log(2\e\kappa)\}$, which gives $O(d^{1+\zeta})$ gradient queries for every fixed $\zeta>0$ at fixed $\kappa$ and $\varepsilon$. The extra factor of $d$ compared with LSI comes from $B$, which enters the bound linearly.

\paragraph{A stationary point does not.}
Under LSI, any point $x_0$ works, and its quality is measured by $G_0$. Under PI, no such guarantee is possible, even when $x_0$ is a strict local minimum.

\begin{theorem}[No uniform bound from an arbitrary stationary point]\label{thm:obstruction}
For fixed $d\ge2$, there is no finite query bound depending only on $d,\beta,\alpha,\varepsilon$ and $\norm{\nabla V(x_0)}$ that guarantees TV error $\varepsilon=1/10$ for all smooth PI targets from an arbitrary supplied reference point. This holds for randomized adaptive exact value-and-gradient oracle algorithms with unrestricted computation between queries. The reference point can be a strict local minimum with Hessian $I_d$.
\end{theorem}

\begin{figure}[t]
\centering
\begin{tikzpicture}[x=0.95cm,y=0.95cm,font=\small]
\fill[black!7] (-1.6,-2.3) rectangle (6,2.6);
\draw[-{Stealth}] (-1.7,-2.3) -- (9.6,-2.3) node[below] {$x$};
\draw[very thick,blue!70!black] (-1.4,2.3) to[out=-60,in=180] (0,0.6) to[out=0,in=180] (0.9,1.0) to[out=0,in=154.5] (2,0.6) -- (6.2,-1.4) to[out=-25.5,in=180] (7,-1.75) to[out=0,in=-137] (7.8,-1.4) -- (9.2,-0.1) node[above left] {$V_R$};
\draw (0,-2.25) -- (0,-2.35) node[below] {$0$};
\draw (6,-2.25) -- (6,-2.35) node[below] {$R-1$};
\draw (7,-2.25) -- (7,-2.35) node[below] {$R$};
\fill (0,0.6) circle (1.6pt);
\node[align=center] at (0,1.55) {strict local\\minimum};
\node[align=center] at (3.7,1.9) {$V_R$ and $V_R'$ do not\\depend on $R$ here};
\node[align=center] (mass) at (7.8,1.2) {most of the\\mass of $\pi_R$};
\draw[-{Stealth}] (mass) -- (7,-1.55);
\end{tikzpicture}
\caption{Schematic potential behind \cref{thm:obstruction}, not to scale. Every $V_R$ has a strict local minimum at the reference point $0$, but most of the mass of $\pi_R\propto\e^{-V_R}$ lies near the remote location $R$. Oracle answers in the shaded region $x\le R-1$ are the same for every $R$, so an algorithm with a fixed query budget cannot locate the mass when $R$ is large.}\label{fig:obstruction}
\end{figure}
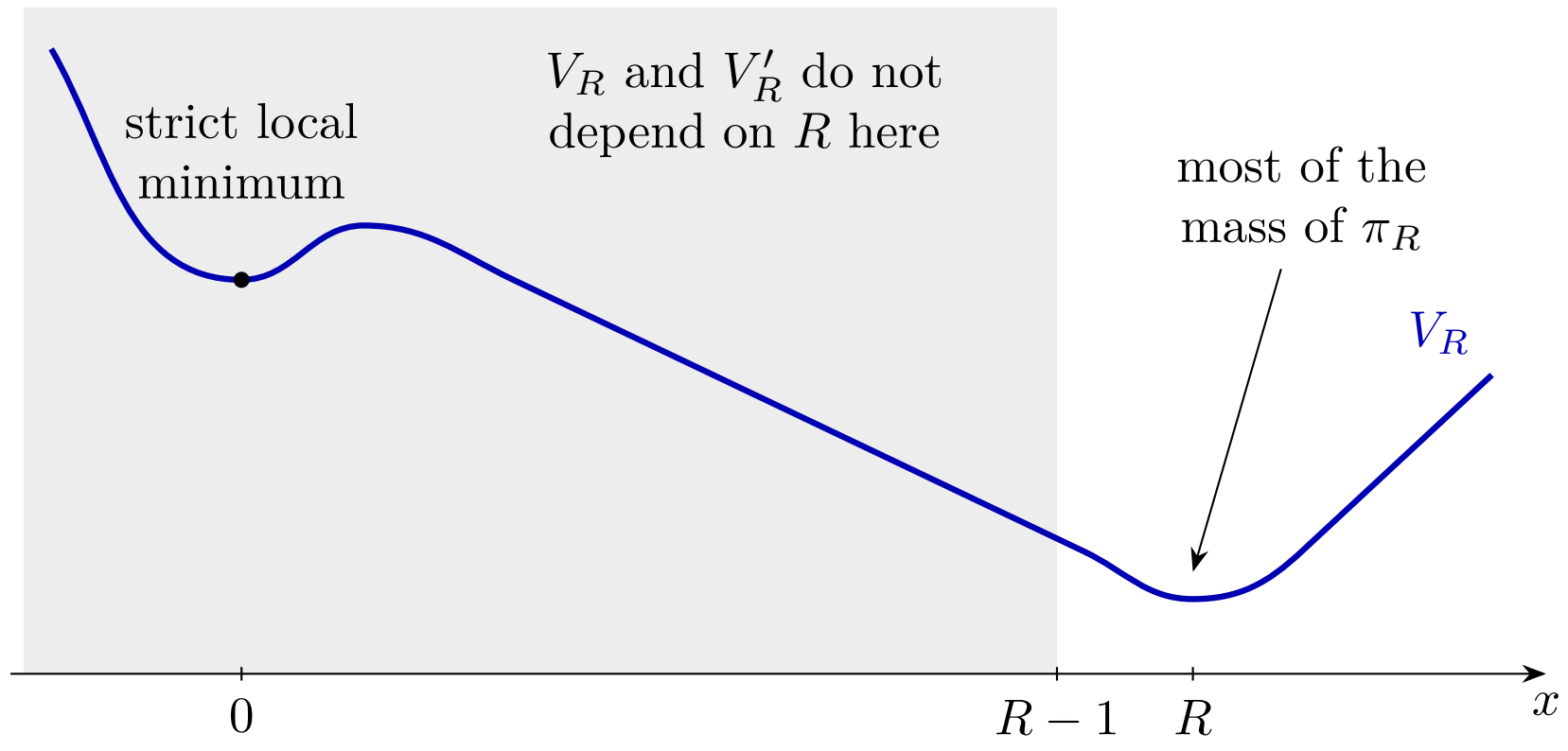

The proof uses a one-dimensional family of targets $\pi_R\propto\e^{-V_R}$, sketched in \cref{fig:obstruction}. Each $\pi_R$ is a smoothed Laplace distribution centered at a remote point $R$, plus a small bounded bump near the origin that creates a strict local minimum there. The bump changes the potential by a bounded amount, so all $\pi_R$ are uniformly smooth and satisfy PI with a common constant. The key property is that, to the left of $R-1$, the potential and its gradient are the same for every $R$. An algorithm with a fixed query budget, started at the origin, therefore receives the same answers for every sufficiently large $R$, so its queries and output stay in a bounded region that does not depend on $R$. Taking $R$ large, the output misses the mass of $\pi_R$. Appending Gaussian coordinates extends the example to any $d\ge2$. \cref{app:obstruction} gives the construction and the argument for randomized adaptive algorithms.

This lower bound concerns the information in the initialization, not the algorithm: it holds even with access to function values, and it is consistent with \cref{thm:pi}, which charges the initialization explicitly.

\section{Conclusions and Discussions}\label{sec:conc}

In this paper, we showed that for smooth targets satisfying LSI or PI, the proximal sampler can be implemented with a number of gradient queries that grows more slowly than any power of the dimension at fixed condition number and accuracy. The implementation runs the observation chain with the noisy local sampler of \cite{openai2026}, controls the accumulated error by comparison with the exact chain, and removes the observation noise once at the end with a decoder. Under LSI, any supplied point can serve for initialization; under PI, the guarantee requires an initialization with a known bound on its R\'enyi divergence, and a lower bound shows that a stationary point alone does not suffice.

Several questions remain open. Our dependence on accuracy is a small power $\varepsilon^{-\nu}$, and it is unknown whether an arbitrarily small dimension exponent can be combined with polylogarithmic dependence on $\varepsilon^{-1}$. Under PI, it would be useful to find initializations that need less information than a global minimizer, or that have a smaller divergence bound $B$, which enters our bound linearly. Finally, intermediate functional inequalities such as the Lata\l a--Oleszkiewicz inequalities \cite{latala2000between} may fit the same framework, given suitable convergence and gradient estimates.

\paragraph{AI usage.}
ChatGPT and Codex assisted with mathematical exploration and proof development, manuscript drafting and revision, literature searches, and generation of code and computational checks. The authors are responsible for the mathematical claims, citations, and final presentation.

\bibliographystyle{alpha}
\bibliography{references}

\clearpage
\appendix
\crefalias{section}{appendix}
\crefalias{subsection}{appendix}

\section{Related Work}
Functional inequalities provide a way to analyze sampling beyond global log-concavity. \cite{vempala2019rapid} proved KL convergence guarantees for the unadjusted Langevin algorithm under LSI and smoothness, without convexity. \cite{chewi2022analysis} developed R\'enyi-divergence guarantees spanning PI and LSI. The proximal analysis in \cite{chen2022improved} separates global convergence from conditional implementation: exact proximal updates contract under the functional inequality, while the Gaussian quadratic makes each conditional strongly log-concave when the step size satisfies $\tau<1/\beta$.
Our analysis uses this separation to control implementation errors along exact histories.

Several approaches implement these conditionals. \cite{gopi2022} developed rejection sampling for Gaussian-regularized Lipschitz potentials in private convex optimization. \cite{liangchen2022} treated weakly smooth, nonsmooth, and composite potentials through rejection-based proximal sampling. The approximate rejection sampler in \cite{fan2023improved} improves dimension dependence through Gaussian concentration. A different approach combines underdamped Langevin warm starts with MALA \cite{altschuler2024faster}, yielding high-accuracy conditional sampling and proximal guarantees beyond log-concavity. Our implementation instead uses the Gaussian-convolved output in \cite{openai2026} to approximate the observation transition directly; it recovers a position sample only through terminal decoding.

Acceleration offers another direction within the proximal framework. The Proximal Bouncy Particle Sampler \cite{chen2026accelerated} combines Gaussian augmentation with local particle dynamics and obtains $\widetilde O(\sqrt\kappa\,d^{1/4})$ expected gradient queries, with polylogarithmic inverse-accuracy dependence, for smooth strongly log-concave targets from an $O(1)$ order-two R\'enyi start. Its curvature and initialization assumptions differ from the nonconvex LSI and PI settings considered here.

Gaussian cloud correction for Picard HMC \cite[Thm.~1.1]{chen2026highpicardhmc} also handles smooth nonconvex LSI targets: it uses $\widetilde O(\kappa d^{1/5})$ expected gradient queries with polylogarithmic inverse-accuracy dependence, assuming $G_0\le d/\kappa$. Our bounds permit any fixed positive dimension exponent, have deterministic query caps, and charge arbitrary $G_0$, but have weaker accuracy dependence. Neither guarantee uniformly dominates.

\section{Auxiliary estimates and local sampling identities}\label{app:decoder}

\subsection{How the noisy local sampler works}\label{app:local-mechanism}\label{sec:primitive-mechanism}
This appendix makes the mechanism behind \cref{prop:primitive} self-contained at the algorithmic level. The construction has three ingredients: a Gaussian probability flow that reduces sampling to conditional means, a second stationary flow that produces noisy estimates of those means, and a finite recursive implementation of both flows using gradient queries. The quantitative derivative estimates, recursive error analysis, and query bound are proved in \cite[Sections~3--7]{openai2026} and are not repeated here.

\paragraph{Gaussian transport reduces sampling to conditional means.}
For independent $Z\sim\rho^F_{c,r}$ and $G\sim\N(0,I_n)$, consider the interpolation $Y_s=sZ+\sqrt{1-s^2}G$, $0\le s<1$, and define
\[
 m_s(y)=\E[\nabla F(c+rZ)\mid Y_s=y].
\]
Completing the square shows that this is another conditional mean of the same type, at a smaller scale:
\begin{equation}\label{eq:mean-recursion}
 m_s(y)=\E_{Q\sim\rho^F_{c+rsy,\,r\sqrt{1-s^2}}}
 \nabla F(c+rsy+r\sqrt{1-s^2}Q).
\end{equation}
The deterministic probability flow
\begin{equation}\label{eq:probability-flow}
 \frac{\d \overline{Y}_s}{\d s}=-r m_s(\overline Y_s),\qquad
 \overline Y_0\sim\N(0,I_n),
\end{equation}
satisfies $\Law(\overline Y_s)=\Law(Y_s)$. This is equality of marginal laws: the interpolating process $Y_s$ does not itself solve the deterministic flow equation. In particular, stopping at $s=1/\sqrt2$ and rescaling by $\sqrt2$ gives the law of $Z+G$. Gaussian conditioning makes the velocity smooth even though $F$ has only two derivatives. The analysis in \cite{openai2026} controls its higher derivatives with dimension dependence compatible with high-order approximation.

\paragraph{Conditional velocity.}
For the interpolation $Y_s=sZ+\sqrt{1-s^2}G$ above, the conditional density gives
\[
 \E[Z\mid Y_s=y]=sy-(1-s^2)r m_s(y)
\]
by integration by parts. Since the derivative of the interpolation is $(Z-sY_s)/(1-s^2)$, its conditional expectation given $Y_s=y$ is $-r m_s(y)$. Differentiating the expectation of a compactly supported smooth test function therefore gives the continuity equation with that velocity. The globally Lipschitz flow in \cref{eq:probability-flow} has the same marginals; \cite[Proposition~3.3]{openai2026} establishes the required regularity under the stated $C^2$ assumption. Completing the conditional square proves \cref{eq:mean-recursion}.

\paragraph{Finite recursion replaces the unknown means by gradient queries.}
The flow is discretized by polynomial quadrature and a fixed number of Picard iterations. Here a Picard iteration substitutes the preceding trajectory approximation into the velocity and integrates; repeated substitution gives a finite expression involving conditional means at other centers. These means are not available as oracles. The construction therefore introduces a second algorithm targeting
\[
 \E_{\rho^F_{c,r}}\nabla F(c+rZ)+\sigma G,
\]
for a prescribed $\sigma>0$ and independent standard Gaussian $G$, using a stationary flow with an auxiliary Gaussian momentum. Discretizing that flow produces further conditional-mean expressions. The sampling and mean algorithms call each other with prescribed error orders and perturbation scales until simple gradient-based approximations suffice. For example, a terminal sampling call has the form
\[
 G^{(0)}-r\nabla F(c+rG^{(0)})+\frac{\sqrt3\sigma_{\rm aux}}{2}G^{(1)}+\frac{\sigma_{\rm aux}}2G^{(2)},
\]
where $\sigma_{\rm aux}>0$ is the requested Gaussian noise standard deviation, distinct from the proximal step size $h$ and the initial density $\eta_0$. With independent standard Gaussian vectors, this approximates $\rho^F_{c,r}*\N(0,\sigma_{\rm aux}^2I_n)$ at a terminal accuracy level. Nonterminal calls use the higher-order flow formulas. The recursion depth is fixed before the quadrature mesh is selected. A sufficiently small mesh exponent, followed by a sufficiently high polynomial approximation order, makes the total query exponent as small as prescribed. Gaussian derivative estimates control the dimension cost of raising that order. The entire finite tree satisfies the deterministic query cap in \cref{prop:primitive}, with constants depending on $\gamma,K,\omega$.

\paragraph{Absorbing a noisy center.}
Suppose a seeded mean algorithm $Q(c;g)$, with $M$ vector-valued seed slots, has the deterministic translation identity
\[
 Q(c+\Delta;g+v\otimes\Delta/r)=Q(c;g),
\]
and preserves every physical gradient-query location in that identity. Here $v\in\R^M$ is fixed, $L>0$, and $\norm v\le L$. Put
\[
 b=r/(2L),\qquad P=(I_M-vv^\top/(4L^2))^{1/2}.
\]
For independent standard Gaussian seed slots $G$ and a standard Gaussian vector $Z$, the call $Q(c+bZ;(P\otimes I_n)G)$ has exactly the law of $Q(c;G)$. Indeed, the translation identity rewrites its seed array at center $c$ as $PG-v\otimes Z/(2L)$, with the tensor identity suppressed. This array is centered Gaussian with slot covariance
\[
 PP^\top+vv^\top/(4L^2)=I_M.
\]
Its slots are therefore independent standard Gaussians. The translation property is maintained recursively; the covariance identity itself is exact. Approximation errors in supplied centers are handled separately by fixed-seed stability, not by assuming the actual centers already have the ideal law.

\paragraph{Reserved smoothing noise.}
Let $W_2(\mu,\eta)$ be the Wasserstein-2 distance, the infimum of $(\E\norm{X-X'}^2)^{1/2}$ over couplings of $X\sim\mu$ and $X'\sim\eta$. The noisy local sampler reserves independent Gaussian noise until all queries are complete. More precisely, the output has the form $U_{\rm pre}+G/2$, where $G$ is independent of $U_{\rm pre}$ and all queries, and
\begin{equation}\label{eq:primitive-reserve}
 W_2(\Law(U_{\rm pre}),
 \rho^F_{c,r}*\N(0,\tfrac34 I_n))
 \le A_n\delta^K(\sqrt n+\norm{\nabla F(c)}).
\end{equation}

For any two laws with finite second moments and any $\sigma>0$, a coupling and the Gaussian shift bound give
\[
 \TV\bigl(\mu*\N(0,\sigma^2I_n),\eta*\N(0,\sigma^2I_n)\bigr)
 \le\frac{W_2(\mu,\eta)}{2\sigma}.
\]
Apply this with $\sigma=1/2$ and the pre-reserve laws in \cref{eq:primitive-reserve}. Adding the independent reserve raises the Gaussian covariance from $3I_n/4$ to $I_n$, yielding the TV guarantee.

\subsection{Normalization and Gaussian padding}

\begin{lemma}[Coordinate and oracle accounting]\label{lem:normalization}
Let $T_a,H_a,\Vpad,\Pi$ be as in \crefrange{eq:coordinate-maps}{eq:padded}, with $n\ge d$. Then $\nabla \Vpad$ is 1-Lipschitz, and $\alpha$-LSI (respectively PI) for $\pi$ implies $1/\kappa$-LSI (respectively PI) for $\Pi$. For any law $\lambda$ on $\R^d$, set $\overline\lambda=(T_a)_\#\lambda\otimes\N(0,I_{n-d})$. Its KL and $D_2$ divergences from $\Pi$ equal those of $\lambda$ from $\pi$. Moreover, $H_{a\#}\Pi=\pi$ and
\[
 \TV(H_{a\#}\eta,\pi)\le\TV(\eta,\Pi)
\]
for every law $\eta$ on $\R^n$. One gradient query to $\Vpad$ costs one gradient query to $V$.
\end{lemma}
\begin{proof}
The normalized Hessian is $\beta^{-1}\nabla^2 V$, and the padding block is the identity. A change of variables in either functional inequality gives constant $\alpha/\beta=1/\kappa$ for $(T_a)_\#\pi$. Since $\alpha\le\beta$, tensorization with the standard Gaussian retains this constant. Under the invertible map $T_a$, density ratios are carried along unchanged. Adding the same independent Gaussian to both laws leaves their KL and $D_2$ integrals unchanged as well. Projection and inverse rescaling give $H_{a\#}\Pi=\pi$, and data processing gives the TV bound. Finally, at $w=(z,\xi)$,
\[
 \nabla \Vpad(w)=\bigl(\beta^{-1/2}\nabla V(a+z/\sqrt\beta),\xi\bigr),
\]
so the gradient of $\Vpad$ follows from a single gradient query at $a+z/\sqrt\beta$.
\end{proof}

\subsection{Stationary score control}

\begin{lemma}[Stationary score identity]\label{lem:score}
If $\Vpad\in C^2(\R^n)$ has 1-Lipschitz gradient and $\e^{-\Vpad}$ is integrable, then
\begin{equation}\label{eq:score}
 \E_\Pi\norm{\nabla \Vpad}^2=\E_\Pi\Delta \Vpad\le n.
\end{equation}
\end{lemma}
\begin{proof}
Choose smooth compactly supported cutoffs $\chi_R$ tending pointwise to one, with $0\le\chi_R\le1$ and $\norm{\nabla\chi_R}\le C/R$. Integration by parts gives
\[
 I_R:=\E_\Pi\chi_R^2\norm{\nabla \Vpad}^2
 =\E_\Pi\chi_R^2\Delta \Vpad+2\E_\Pi\chi_R\nabla\chi_R\cdot\nabla \Vpad.
\]
Since $|\Delta \Vpad|\le n$, Young's inequality gives $I_R\le n+I_R/2+C/R^2$. Fatou's lemma proves square integrability of the score. The cross term then tends to zero by Cauchy--Schwarz. Dominated convergence for the bounded Hessian trace and the integrable squared score proves \cref{eq:score}.
\end{proof}

\subsection{Local estimates used by the decoder}

We also use the following two local consequences from \cite[Lemma~2.2, Eq.~(2.6), and Lemma~8.4]{openai2026}. For $\Lip(\nabla F)\le\delta$ and $\delta r^2\le1/2$, if $Z\sim\rho^F_{c,r}$ and $G$ is an independent standard Gaussian, then
\begin{equation}\label{eq:local-moment}
 (\E\norm{Z+G}^2)^{1/2}\le C(\sqrt n+r\norm{\nabla F(c)}).
\end{equation}
If $v_*$ is the mode of $\rho^F_{c,r}$, then
\begin{equation}\label{eq:local-gaussian}
 \TV(\rho^F_{c,r},\N(v_*,I_n))\le C\delta r^2\sqrt n.
\end{equation}
These are local strongly log-concave estimates and remain valid when the global target is non-log-concave.

\subsection{Proof of the decoder guarantee}

\begin{proof}
Let $Y=X+\sqrt hG$ with $X\sim\Pi$ and independent standard Gaussian $G$. By \cref{lem:score} and Lipschitz continuity,
\[
 \E\norm{\nabla \Vpad(Y)}^2\le2n+2hn\le3n.
\]
\emph{Exact-history moments.} Define a separate exact process by $c_0^\star=0$, $r_j=2^{-j/2}$, and
\[
 U_j^\star\mid(Y,c_j^\star)\sim
 \rho^{F_Y}_{c_j^\star,r_j}*\N(0,I_n),\qquad
 c_{j+1}^\star=c_j^\star+r_jU_j^\star/2.
\]
These are exact draws, not calls to $\primitive$. Put $b_j=(\E\norm{g_Y(c_j^\star)}^2)^{1/2}$, where expectation includes $Y\sim\Pi_h$ and the exact history. Initially $b_0\le\sqrt{3hn}$. Conditional on $Y,c_j^\star$, \cref{eq:local-moment} applies directly:
\[
 (\E[\norm{U_j^\star}^2\mid Y,c_j^\star])^{1/2}
 \le C(\sqrt n+r_j\norm{g_Y(c_j^\star)}).
\]
The exact center update, $\Lip(g_Y)\le\delta$, and the $L^2$ triangle inequality imply
\[
 b_{j+1}\le(1+C\delta r_j^2)b_j+C\delta r_j\sqrt n.
\]
Both $\sum_jr_j$ and $\sum_jr_j^2$ are bounded by absolute constants, so iteration gives $b_j\le C\sqrt n$ uniformly in $j$.

\emph{Implementation error.} Let $\mathcal H_j$ be the output law, averaged over $Y\sim\Pi_h$, obtained with the first $j$ observations exact and the remaining $J-j$ observations implemented by $\primitive$. Every hybrid uses the same terminal Gaussian rule \crefrange{eq:fixed-point}{eq:decoded-output}. Thus $\mathcal H_0=\Pi_h\widehat R_h$, while $\mathcal H_J$ has an entirely exact observation history. Consecutive hybrids share the exact prefix $(Y,c_j^\star)$. Conditional on this prefix, \cref{eq:source-error} bounds the differing observation; all subsequent computation is a common Markov kernel. Hence
\[
 \TV(\mathcal H_j,\mathcal H_{j+1})
 \le A_n\delta^K\E[\sqrt n+\norm{g_Y(c_j^\star)}]
 \le CA_n\delta^K\sqrt n.
\]
Summing yields
\begin{equation}
 \TV(\mathcal H_0,\mathcal H_J)\le CJA_n\delta^K\sqrt n=:E_{\rm descent}.
\end{equation}
No moment bound for the implemented histories is used.

\emph{Terminal approximation.} Under the exact history, let $v_*^\star$ be the mode of $\rho^{F_Y}_{c_J^\star,r_J}$ and let $v^{(m),\star}$ denote the fixed-point iterate at that center. The mode satisfies $v_*^\star=-r_Jg_Y(c_J^\star+r_Jv_*^\star)$. This map is $l$-contractive, where $l=\delta r_J^2\le\delta n^{-4}\le1/2$, so
\[
 \norm{v^{(m),\star}-v_*^\star}\le l^m\norm{v_*^\star}
 \le\frac{l^m r_J\norm{g_Y(c_J^\star)}}{1-l}.
\]
By \cref{eq:local-gaussian}, the Gaussian shift inequality, and $b_J\le C\sqrt n$, the averaged error in replacing this last conditional by the terminal Gaussian is at most
\[
 C\delta r_J^2\sqrt n+
 C\frac{(\delta r_J^2)^m r_J\sqrt n}{1-\delta r_J^2}
 \le C\delta n^{-7/2}=:E_{\rm base}.
\]
The last inequality uses $m\ge1$ and $r_J\le1$. Conditional affine reconstruction contracts TV. With an exact terminal draw, repeated use of \cref{eq:halving} reconstructs $R_h(\,\cdot\mid Y)$, whose average under $\Pi_h$ is $\Pi$. Thus $\TV(\mathcal H_J,\Pi)\le E_{\rm base}$, and the triangle inequality proves \cref{eq:decoder-error}. The cost is $J$ calls to $\primitive$ and $m$ gradient iterations.
\end{proof}

\subsection{The bounded-perturbation example}\label{app:example}
The LSI comparison used in \cref{sec:lsi-proof} follows from
\[
 \Ent_\mu(f^2)=\inf_{a>0}\E_\mu[f^2\log(f^2/a)-f^2+a].
\]
The integrand is nonnegative. If $m\le\mu/\Gamma\le M$ and $\Gamma$ is standard Gaussian, its LSI gives $\Ent_\mu(f^2)\le M\Ent_\Gamma(f^2)\le2(M/m)\E_\mu\norm{\nabla f}^2$. For the perturbation $2\cos x_1$, $M/m\le\e^4$. Its Hessian eigenvalues lie between $-1$ and $3$, proving the example's other claims. Direct Gaussian rejection uses acceptance $\e^{-2-2\cos x_1}$; this additional structure explains why the example is not a complexity-separation witness.

\section{Proofs for the LSI analysis}\label{app:lsi-proofs}
Throughout, $\Pi$ has LSI constant $1/\kappa$ (\cref{lem:normalization}), and since $a=x_0$, we have $\norm{\nabla \Vpad(0)}^2=G_0$.

\begin{lemma}[Gaussian initial entropy]\label{lem:gaussian-kl}
For $\lambda_0=\N(0,I_n)$,
\begin{equation}
 \KL(\lambda_0\Vert\Pi)\le\kappa(G_0+4n).
\end{equation}
\end{lemma}
\begin{proof}
For $Z\sim\N(0,I_n)$, the relative score is $\nabla\log(\lambda_0/\Pi)(Z)=\nabla \Vpad(Z)-Z$. Since $\nabla\Vpad$ is 1-Lipschitz,
\[
 \norm{\nabla \Vpad(Z)-Z}\le\norm{\nabla \Vpad(0)}+2\norm Z,
\]
so $\E_{\lambda_0}\norm{\nabla\log(\lambda_0/\Pi)}^2\le2G_0+8n$. Apply \cref{eq:lsi}, with constant $1/\kappa$, to $f=(\lambda_0/\Pi)^{1/2}$. Then $\Ent_\Pi(f^2)=\KL(\lambda_0\Vert\Pi)$ and $4\E_\Pi\norm{\nabla f}^2=\E_{\lambda_0}\norm{\nabla\log(\lambda_0/\Pi)}^2$, so
\[
 \KL(\lambda_0\Vert\Pi)\le\frac{2\kappa}{4}(2G_0+8n)=\kappa(G_0+4n).
\]
The last expectation is finite, so $f$ is an admissible test function; alternatively, apply \cref{eq:lsi} to truncations and pass to the limit.
\end{proof}

The same argument without padding ($n=d$) shows that the initialization in the original coordinates satisfies $\KL(\N(x_0,\beta^{-1}I_d)\Vert\pi)\le\kappa(G_0+4d)$.

\begin{lemma}[Gradient moments from relative entropy]\label{lem:lsi-moments}
If $\KL(\lambda\Vert\Pi)\le b$, then
\begin{equation}
 \E_\lambda\norm{\nabla \Vpad(X)}^2\le2n+4\kappa b.
\end{equation}
If $X\sim\lambda$, $G\sim\N(0,I_n)$ is independent, and $Y=X+\sqrt hG$ with $h\le1/2$, then
\begin{equation}
 \E\norm{\nabla \Vpad(Y)}^2\le5n+8\kappa b.
\end{equation}
\end{lemma}
\begin{proof}
Since $\Pi$ satisfies LSI with constant $1/\kappa$, it satisfies Talagrand's transport inequality \cite{otto2000generalization}
\begin{equation}\label{eq:t2}
 W_2^2(\lambda,\Pi)\le2\kappa\KL(\lambda\Vert\Pi).
\end{equation}
Couple $X\sim\lambda$ to $X_*\sim\Pi$. Since $\nabla\Vpad$ is 1-Lipschitz,
\[
 \E\norm{\nabla \Vpad(X)}^2
 \le2\E\norm{\nabla \Vpad(X_*)}^2+2\E\norm{X-X_*}^2.
\]
Taking the infimum over couplings, the first term is at most $2n$ by \cref{lem:score}, and the second is at most $2W_2^2(\lambda,\Pi)\le4\kappa b$ by \cref{eq:t2}. For $Y$, the same argument gives
\[
 \E\norm{\nabla \Vpad(Y)}^2\le2\E\norm{\nabla \Vpad(X)}^2+2h\E\norm G^2\le4n+8\kappa b+2hn\le5n+8\kappa b.\qedhere
\]
\end{proof}

\begin{proof}[Proof of \cref{prop:lsi-inputs}]
Set $B_0=5\kappa n$. Since $G_0\le n$, \cref{lem:gaussian-kl} gives $\KL(\lambda_0\Vert\Pi)\le B_0$. By \cref{prop:ideal-convergence}, applied to $\Pi$ with LSI constant $1/\kappa$ and step size $h$,
\[
 \KL(\lambda_j\Vert\Pi)\le(1+h/\kappa)^{-2j}B_0\qquad\text{for all }j\ge0.
\]
Choose
\begin{equation}\label{eq:N-lsi}
 N=\ceil{\frac{\log(8B_0/\varepsilon^2)}{2\log(1+h/\kappa)}}
 \le C\kappa\delta^{-1}\log n,
\end{equation}
where the inequality uses $\log(1+h/\kappa)\ge h/(2\kappa)$ and $\kappa,\varepsilon^{-1}\le n$. Then $\KL(\lambda_N\Vert\Pi)\le\varepsilon^2/8$. Since $\mu_N=\lambda_NA_h$ and $\Pi_h=\Pi A_h$, data processing and Pinsker's inequality give $\varepsilon_{\rm mix}=\TV(\mu_N,\Pi_h)\le\TV(\lambda_N,\Pi)\le\varepsilon/4$. For the score budget, every $\lambda_j$ has divergence at most $B_0$, and $\mu_j$ is the law of $X+\sqrt hG$ with $X\sim\lambda_j$. Hence \cref{lem:lsi-moments} gives $\E_{\mu_j}\norm{\nabla \Vpad(Y)}^2\le5n+40\kappa^2n\le45\kappa^2n$, and by Cauchy--Schwarz each summand of $M_N$ is at most $C\kappa\sqrt n$. Therefore
\begin{equation}\label{eq:lsi-outer}
 M_N\le C N\kappa\sqrt n,\qquad
 A_n\delta^K M_N\le C\kappa^2 A_n\delta^{K-1}\sqrt n\log n.\qedhere
\end{equation}
\end{proof}

\begin{proof}[Proof of \cref{thm:lsi}]
By \cref{eq:query-general}, $\delta^{-1}=n^\gamma$, and \cref{eq:N-lsi}, the algorithm makes at most $C\kappa A_n(\log n)\,n^{\gamma(1+\omega)}$ queries after computing $G_0$. With $n$ as in \cref{eq:n-lsi} and the parameter choices of \cref{app:parameters}, the implementation error \cref{eq:lsi-outer} and the decoder error \cref{eq:decoder-error} are each at most $\varepsilon/4$, so \cref{prop:transfer,prop:lsi-inputs} bound the padded output error by $\varepsilon$, and the query count is of the form \cref{eq:lsi-main}. Finally, for the law $\widehat\pi$ returned by \cref{alg:proximal}, \cref{lem:normalization} transfers the error back to the original target:
\begin{equation}\label{eq:original-tv}
 \TV(\widehat\pi,\pi)
 =\TV\bigl(H_{a\#}(\widehat\mu_N\widehat R_h),H_{a\#}\Pi\bigr)
 \le\TV(\widehat\mu_N\widehat R_h,\Pi)\le\varepsilon.\qedhere
\end{equation}
\end{proof}

\section{Proofs for the PI analysis}\label{app:pi-proofs}
Throughout, $\Pi$ has PI constant $1/\kappa$ (\cref{lem:normalization}).

\begin{lemma}[A PI exponential-moment bound]\label{lem:pi-mgf}
If $\Pi$ satisfies $1/\kappa$-PI and $F$ is integrable and 1-Lipschitz, then
\begin{equation}
 \log\E_\Pi\e^{(F-\E_\Pi F)/\sqrt\kappa}\le1.
\end{equation}
\end{lemma}
\begin{proof}
For bounded $F$, write $M(s)=\E\e^{sF}$. Applying PI to $\e^{sF/2}$ gives
\[
 M(s)-M(s/2)^2\le\frac{\kappa s^2}{4}M(s).
\]
Set $s=\kappa^{-1/2}$ and iterate the corresponding logarithmic inequality at $s,s/2,s/4,\ldots$. After subtracting the centered linear term,
\[
 \log M(s)-s\E F
 \le\sum_{j=0}^\infty2^j\left[-\log\left(1-\frac{1}{4\cdot4^j}\right)\right]
 \le\frac12\sum_{j=0}^\infty2^{-j}=1.
\]
The omitted limiting centered term is zero because $F$ is bounded. The second inequality uses $-\log(1-u)\le2u$ for $u\le1/4$. Apply this result to Lipschitz truncations of $F$. Their means converge by integrability, and Fatou's lemma gives the claim. Weak-gradient approximation justifies the test functions when $F$ is only Lipschitz.
\end{proof}

\begin{lemma}[PI change of measure for the score]\label{lem:pi-score}
If $\KL(\lambda\Vert\Pi)\le B$, then
\begin{equation}\label{eq:pi-score}
 \E_\lambda\norm{\nabla \Vpad(X)}\le\sqrt n+\sqrt\kappa(B+1).
\end{equation}
For $X\sim\lambda$ and $Y=X+\sqrt hG$ with independent $G\sim\N(0,I_n)$,
\begin{equation}
 \E\norm{\nabla \Vpad(Y)}\le(1+\sqrt h)\sqrt n+\sqrt\kappa(B+1).
\end{equation}
\end{lemma}
\begin{proof}
The function $F=\norm{\nabla \Vpad}$ is 1-Lipschitz and $\E_\Pi F\le\sqrt n$ by \cref{lem:score}. The entropy variational inequality and \cref{lem:pi-mgf}, with $s=\kappa^{-1/2}$, give
\[
 s\E_\lambda F\le\KL(\lambda\Vert\Pi)+\log\E_\Pi \e^{sF}
 \le B+s\E_\Pi F+1.
\]
Use nonnegative truncations of $F$ first if its integrability under $\lambda$ has not yet been established. This proves \cref{eq:pi-score}. The Gaussian observation adds at most $\sqrt h\E\norm G\le\sqrt{hn}$ by Lipschitz continuity.
\end{proof}

\begin{proof}[Proof of \cref{prop:pi-inputs}]
By \cref{lem:normalization}, normalizing the initialization and appending independent Gaussian coordinates preserves its $D_2$ divergence, so $\chi^2(\lambda_0\Vert\Pi)\le\e^B-1$ and $\KL(\lambda_0\Vert\Pi)\le B$. By \cref{prop:ideal-convergence}, applied to $\Pi$ with PI constant $1/\kappa$ and step size $h$, together with data processing under $A_h$,
\[
 \chi^2(\mu_j\Vert\Pi_h)\le(\e^B-1)(1+h/\kappa)^{-2j},
 \qquad \KL(\lambda_j\Vert\Pi)\le B.
\]
Choose
\begin{equation}\label{eq:N-pi}
 N=\ceil{\frac{B+2\log(2/\varepsilon)}{2\log(1+h/\kappa)}}
 \le C\kappa H\delta^{-1}.
\end{equation}
Then $\chi^2(\mu_N\Vert\Pi_h)\le\varepsilon^2/4$, and $4\TV^2\le\chi^2$ gives $\varepsilon_{\rm mix}\le\varepsilon/4$. For the score budget, $\mu_j$ is the law of $X+\sqrt hG$ with $X\sim\lambda_j$, and $\sqrt\kappa(B+1)\le\sqrt n$. Hence \cref{lem:pi-score} gives $\E_{\mu_j}\norm{\nabla \Vpad(Y)}\le3\sqrt n$, and therefore
\[
 M_N\le C N\sqrt n,\qquad
 A_n\delta^K M_N\le C\kappa H A_n\delta^{K-1}\sqrt n.\qedhere
\]
\end{proof}

\begin{proof}[Proof of \cref{thm:pi}]
The decoder bound is the same as under LSI. With $n$ as in \cref{eq:n-pi} and the parameter choices of \cref{app:parameters}, the implementation and decoder errors are each at most $\varepsilon/4$, so \cref{prop:transfer,prop:pi-inputs} bound the padded output error by $\varepsilon$, and \cref{eq:pi-query-padded} gives a query count of the form \cref{eq:pi-main}. Applying \cref{eq:original-tv} with $a=0$ proves the original-space guarantee.
\end{proof}

\section{Constants and Gaussian initialization under PI}\label{app:pi-details}
\subsection{Functional-inequality constants and smoothness}

\begin{lemma}\label{lem:alpha-beta}
If $\nabla V$ is $\beta$-Lipschitz and $\pi\propto\e^{-V}$ satisfies $\alpha$-PI, then $\alpha\le\beta$.
\end{lemma}
\begin{proof}
Fix a unit vector $e\in\R^d$, let $X\sim\pi$, and write $m=\E\langle e,X\rangle$.

Since $\nabla\e^{-V}=-\nabla V\,\e^{-V}$, every $C^1$ function $g\colon\R^d\to\R$ such that $g$ and $\nabla g$ grow at most linearly satisfies
\begin{equation}\label{eq:alpha-beta-ibp}
 \E\langle e,\nabla g(X)\rangle=\E[g(X)\langle e,\nabla V(X)\rangle],
\end{equation}
by applying the divergence theorem to the function $g\,\e^{-V}e$.

Taking $g\equiv1$ in \cref{eq:alpha-beta-ibp} gives $\E\langle e,\nabla V(X)\rangle=0$. Taking $g(x)=\langle e,x\rangle$ gives $\E[\langle e,X\rangle\langle e,\nabla V(X)\rangle]=\norm e^2=1$; subtracting $m\,\E\langle e,\nabla V(X)\rangle=0$ yields
\[
 \E[(\langle e,X\rangle-m)\langle e,\nabla V(X)\rangle]=1.
\]
Taking $g(x)=\langle e,\nabla V(x)\rangle$ and using $\nabla^2V\preceq\beta I_d$ gives
\[
 \E[\langle e,\nabla V(X)\rangle^2]=\E\langle e,\nabla^2V(X)\,e\rangle\le\beta.
\]

By the Cauchy--Schwarz inequality,
\[
 1=(\E[(\langle e,X\rangle-m)\langle e,\nabla V(X)\rangle])^2
 \le\Var\langle e,X\rangle \E[\langle e,\nabla V(X)\rangle^2]
 \le\beta\Var\langle e,X\rangle
 \implies
 \Var\langle e,X\rangle\ge1/\beta.
\]
On the other hand, applying \cref{eq:pi} to $f(x)=\langle e,x\rangle$ gives $\Var\langle e,X\rangle\le1/\alpha$.
Combining the two bounds completes the proof.
\end{proof}
\subsection{Gaussian initialization near a global minimizer}\label{app:pi-init}

\begin{proposition}[Gaussian divergence near a global minimizer]\label{prop:gaussian-pi}
Suppose $\pi$ satisfies $\alpha$-PI and a global minimizer $x_*$ of $V$ exists. For a point $x_0$ and a known $A_0\ge V(x_0)-V(x_*)$, let
\[
 q_0=\N(x_0,(2\beta)^{-1}I_d),\qquad G_0=\norm{\nabla V(x_0)}^2/\beta.
\]
Then
\begin{equation}
 D_2(q_0\Vert\pi)\le D_\infty(q_0\Vert\pi)
 \le A_0+G_0/2+\frac d2\log(2\e\kappa)
 \le2A_0+\frac d2\log(2\e\kappa).
\end{equation}
In particular, at $x_0=x_*$ the first two terms vanish.
\end{proposition}
\begin{proof}
For this proof, write $Z_V=\int_{\R^d}\e^{-V(x)}\,\d x$. First, PI implies $\operatorname{Cov}_\pi X\preceq\alpha^{-1}I_d$. One can justify finite second moments before applying PI to a coordinate as follows. Truncate a coordinate centered at a median to $[-R,R]$. The truncation is 1-Lipschitz and retains zero as a median, so its variance is at most $\alpha^{-1}$ and the square of its mean is at most twice that variance. Fatou's lemma gives a finite second moment about the median. Applying PI to truncations of arbitrary linear functionals and passing to the limit gives the covariance bound. Smoothness then gives $\E_\pi|V|<\infty$.

The Gaussian maximum-entropy bound, or nonnegativity of KL relative to a Gaussian with covariance $\alpha^{-1}I_d$, implies
\[
 \mathsf h(\pi)\le\frac d2 \log \frac{2\pi\e}\alpha.
\]
Since $\mathsf h(\pi)=\log Z_V+\E_\pi V$ and $\E_\pi V\ge V(x_*)$,
\begin{equation}
 \log Z_V+V(x_*)\le\frac d2 \log \frac{2\pi\e}\alpha.
\end{equation}
For $u=x-x_0$, the smoothness upper bound for $V$ gives
\begin{align*}
 \log\frac{q_0(x)}{\pi(x)}
 &\le\log Z_V+V(x_0)+\frac d2 \log\frac{\beta}\pi
       +\nabla V(x_0)\cdot u-\frac\beta2\norm u^2\\
 &\le A_0+\frac{G_0}{2}+\frac d2\log(2\e\kappa).
\end{align*}
Here $D_\infty(q_0\Vert\pi)=\log\operatorname*{ess\,sup}(q_0/\pi)$. Taking the essential supremum proves the first upper bound. Finally, the descent lemma evaluated at $x_0-\nabla V(x_0)/\beta$ gives $G_0/2\le V(x_0)-V(x_*)\le A_0$.
\end{proof}

\section{The remote-mass construction}\label{app:obstruction}

Define
\[
 U(t)=\begin{cases}
 \frac38+\frac34t^2-\frac18t^4,&|t|\le1,\\
 |t|,&|t|\ge1,
 \end{cases}
 \qquad \phi(x)=(x+x^2/2)\e^{-x^2},
\]
and, for $R\ge2$, set
\begin{equation}
 V_R(x)=U(x-R)-R+\phi(x),\qquad \pi_R(x)\propto\e^{-V_R(x)}.
\end{equation}

\begin{lemma}[Uniform geometry and identical local responses]\label{lem:remote}
The potentials $V_R$ are normalizable, $C^2$, and 20-smooth. They satisfy
\[
 V_R(0)=0,\qquad V_R'(0)=0,\qquad V_R''(0)=1.
\]
Their Gibbs laws have PI constant at least $\alpha_0=(4\e^{19/8})^{-1}$, independently of $R$. For $A\le R$,
\begin{equation}\label{eq:remote-tail}
 \pi_R((-\infty,A])\le\tfrac12\e^{19/8}\e^{A-R}.
\end{equation}
For every $x\le R-1$, the value-and-gradient oracle returns
\begin{equation}\label{eq:common-oracle}
 V_R(x)=-x+\phi(x),\qquad V_R'(x)=-1+\phi'(x),
\end{equation}
independently of $R$.
\end{lemma}
\begin{proof}
The polynomial branch of $U$ matches $|t|$ in value and in its first two derivatives at $\pm1$. It satisfies $0\le U-|\cdot|\le3/8$ and $0\le U''\le3/2$. Also $\phi(0)=0$, $\phi'(0)=\phi''(0)=1$, $\norm\phi_\infty<1$, and
\[
 \phi''(x)=(1-6x-5x^2+4x^3+2x^4)\e^{-x^2}.
\]
Since $\sup_x|x|^j\e^{-x^2}\le1$ for $j=0,\ldots,4$, we have $\norm{\phi''}_\infty\le18$. These facts prove the smoothness, normalizability, and origin claims.

The centered Laplace density $\ell(x)=\e^{-|x|}/2$ has PI constant $1/4$. Indeed, for $g(0)=0$, integration by parts on a half-line gives
\[
 \int_0^\infty g(x)^2\e^{-x}\,\d x
 =2\int_0^\infty g(x)g'(x)\e^{-x}\,\d x
 \le2\left(\int_0^\infty g(x)^2\e^{-x}\,\d x\right)^{1/2}
        \left(\int_0^\infty g'(x)^2\e^{-x}\,\d x\right)^{1/2}.
\]
Apply the resulting factor-four inequality to $g=f-f(0)$ on both half-lines, and use $\Var_\ell f\le\E_\ell(f-f(0))^2$. Bounded smooth functions are justified by cutoff approximation.

Relative to the Laplace density translated by $R$, the potential perturbation is
\[
 b_R(x)=U(x-R)-|x-R|+\phi(x),\qquad \osc b_R\le19/8.
\]
If a density ratio lies between $m$ and $M$, comparison of the variance using its infimum over constants and comparison of the Dirichlet form worsen the PI constant by at most $M/m$. Here $M/m\le\e^{19/8}$, giving PI constant $\alpha_0$. The same density comparison and the translated Laplace left tail give \cref{eq:remote-tail}. When $x\le R-1$, $U(x-R)=R-x$, proving \cref{eq:common-oracle}.
\end{proof}

\subsection{Proof of the lower bound}

\begin{proof}
First work in dimension one. Suppose a total oracle algorithm has a uniform deterministic query cap $q$ and the asserted accuracy on the family in \cref{lem:remote}, with common supplied bounds $\beta=20$, $\alpha=\alpha_0$, and reference point zero.

Run its decision rules against the common response function $V_\infty(x)=-x+\phi(x)$. This is a device for generating transcripts, not a normalizable target. Every finite transcript is consistent with some admissible $V_R$, by choosing $R$ beyond all query locations in that transcript. Thus the rules produce a finite next query or an output, and they cannot make a $(q+1)$st query: such a finite prefix would also occur for an admissible target and violate the uniform cap. Totality on admissible inputs likewise rules out nontermination between queries along a realizable finite prefix.

For a random seed, the at most $q$ query locations and the final output are finitely many finite real random variables. Choose a deterministic $A$ such that, with probability at least $0.99$, all are at most $A$. For $R>A+1$, couple the two runs with the same seed. On that event all oracle replies and the final output agree with the common-response run. The algorithm's output under $V_R$ therefore lies in $(-\infty,A]$ with probability at least $0.99$. Increase $R$ until \cref{eq:remote-tail} is at most $0.01$. Its TV error is at least $0.98$, a contradiction.

For fixed $d\ge2$, use
\[
 \widetilde V_R(x)=V_R(x_1)+\frac12\sum_{i=2}^d x_i^2.
\]
Tensorization preserves the common PI lower bound, smoothness remains at most 20, and the Hessian at the origin is $I_d$. Apply the same transcript argument to the first coordinate of each query and output.
\end{proof}

\section{Parameter selection and deterministic query accounting}\label{app:parameters}
The local-sampler parameters are fixed before the target dimension and accuracy are supplied. A convenient explicit recipe is
\[
 \gamma_{\rm LSI}=\min\{1/8,\nu/4\},\qquad
 \gamma_{\rm PI}=\min\{1/8,\nu/12\},\qquad
 K=\lfloor2+4/\gamma\rfloor+1.
\]
In either regime take $\omega$ as in \cref{eq:omega-choice} below. A fixed integer threshold $n_*(\nu)$ dominates the threshold in \cref{prop:primitive}, ensures $n\ge2$ and $\delta\le1/2$, and absorbs the fixed-order logarithmic factors. Its size is not quantitatively bounded here. We verify below that the choices have strict margins for both error and cost.

\subsection{LSI parameter bounds}
Write $E_{\rm outer}=A_n\delta^K M_N$, and let $E_{\rm descent}$ and $E_{\rm base}$ denote the two decoder contributions bounded in \cref{app:decoder}.

Fix $\nu>0$. Choose $0<\gamma<1/4$ with $2\gamma<\nu$, choose an integer $K>2+4/\gamma$, and set, for example,
\begin{equation}\label{eq:omega-choice}
 \omega=\frac{1}{1000(K+2)}.
\end{equation}
The constraint in \cref{eq:source-gate} holds, and $\gamma(K-1)>4$, $\omega<1$, and $\gamma(1+\omega)<\nu$. Using $\kappa\le n$ in \cref{eq:lsi-outer},
\[
 E_{\rm outer}\le CA_n(\log n)n^{5/2-\gamma(K-1)}=o(n^{-1}).
\]
Also $E_{\rm descent}\le CA_n(\log n)n^{1/2-\gamma K}=o(n^{-1})$ and $E_{\rm base}\le Cn^{-7/2-\gamma}=o(n^{-1})$. The polylogarithmic factors have constants depending only on the fixed local-sampler parameters. Hence a single threshold $n_*(\nu)$ makes each of these three errors at most $1/(4n)$ for all $n\ge n_*(\nu)$.

Now take
\begin{equation}\label{eq:n-lsi}
 n=\max\{d,n_*(\nu),\ceil\kappa,\ceil{\varepsilon^{-1}},\ceil{G_0}\}.
\end{equation}
Then $1/n\le\varepsilon$, and \cref{prop:transfer} proves the accuracy claim. By \cref{eq:query-general} and \cref{eq:N-lsi},
\[
 Q\le1+C\kappa\operatorname{polylog}(n)n^{\gamma(1+\omega)}
 \le C_\nu\kappa n^\nu.
\]
The second inequality absorbs both the initial query and the logarithmic factor by increasing $C_\nu$, using $\kappa n^\nu\ge1$. The dimension choice satisfies $n\le C_\nu(d+\kappa+\varepsilon^{-1}+G_0)$, proving \cref{eq:lsi-main} after projection and inverse rescaling. All loop counts depend only on known parameters after the first reference query. The query cap of $\primitive$ is uniform in all subsequently generated centers, so the resulting query bound is deterministic. \hfill$\square$

\subsection{PI parameter bounds}

Fix $\nu>0$, choose $0<\gamma<1/4$ with $6\gamma<\nu$, choose $K>2+4/\gamma$, and choose $\omega$ by \cref{eq:omega-choice}. Let
\begin{equation}\label{eq:n-pi}
 n=\max\{d,n_*(\nu),\ceil{\kappa(B+1)^2},\ceil{\varepsilon^{-1}}\},
\end{equation}
where the threshold is enlarged below as needed. Then $\kappa\le n$, $B\le\sqrt n$, and $H\le Cn$. Thus
\[
 E_{\rm outer}\le CA_n n^{5/2-\gamma(K-1)}=o(n^{-1}).
\]
The decoder errors are the same as in the LSI proof and are also $o(n^{-1})$. A fixed threshold makes each at most $1/(4n)\le\varepsilon/4$, proving accuracy by \cref{prop:transfer}.

By \cref{eq:query-general},
\begin{equation}\label{eq:pi-query-padded}
 Q-Q_{\rm init}\le C\kappa H\operatorname{polylog}(n)n^{\gamma(1+\omega)}
 \le C_\nu\kappa H n^{\nu/3},
\end{equation}
since $3\gamma(1+\omega)<6\gamma<\nu$. Put $T=d+\kappa+B+\varepsilon^{-1}$. Because $\kappa(B+1)^2\le4T^3$ and the other padding terms are also at most $C_\nu T^3$, we have $n\le C_\nu T^3$. Substitution into \cref{eq:pi-query-padded} proves \cref{eq:pi-main}. As in the LSI case, every invocation is uniformly capped and every loop length is predetermined from supplied parameters. \hfill$\square$

\end{document}